\documentclass[journal]{IEEEtran}
\usepackage{multirow}
\usepackage{epsfig}
\usepackage{amsmath}
\usepackage{amssymb}
\usepackage{amsfonts}
\usepackage{url}
\usepackage{color}
\usepackage{epsfig}
\usepackage{graphicx}
\usepackage{amsfonts,amssymb}

\ifCLASSINFOpdf
\else
\fi
\begin{document}
%
\title{Weakly-Supervised Monocular Depth Estimation with Resolution-Mismatched Data}

\author{Jialei Xu,
Yuanchao Bai,~\IEEEmembership{Member,~IEEE,}
Xianming~Liu,~\IEEEmembership{Member,~IEEE,}
Junjun~Jiang,~\IEEEmembership{Member,~IEEE,}
Xiangyang Ji,~\IEEEmembership{Member,~IEEE}
\thanks{This work was supported by XX.}

\IEEEcompsocitemizethanks{
\IEEEcompsocthanksitem J. Xu, Y. Bai, X. Liu, and J. Jiang  are with the School of Computer Science and Technology, Harbin Institute of Technology, Harbin 150001, China, and also with the Peng Cheng Laboratory, Shenzhen 518052, China  E-mail: 20S003044@stu.hit.edu.cn, \{csxm,jiangjunjun\}@hit.edu.cn, yuanchao.bai@pku.edu.cn.
\IEEEcompsocthanksitem X. Ji is with the Department of Automation, Tsinghua University, Beijing 100084, China.  E-mail: xyji@tsinghua.edu.cn.}

\thanks{}}
\maketitle

\begin{abstract}
Depth estimation from a single image is an active research topic in computer vision. The most accurate approaches are based on fully supervised learning models, which rely on a large amount of dense and high-resolution (HR) ground-truth depth maps. However, in practice, color images are usually captured with much higher resolution than depth maps, leading to the resolution-mismatched effect.
In this paper, we propose a novel weakly-supervised framework to train a monocular depth estimation network to generate HR depth maps with resolution-mismatched supervision, i.e., the inputs are HR color images and the ground-truth are low-resolution (LR) depth maps. The proposed weakly supervised framework is composed of a sharing weight monocular depth estimation network and a depth reconstruction network for distillation. Specifically, for the monocular depth estimation network the input color image is first downsampled to obtain its LR version with the same resolution as the ground-truth depth. Then, both HR and LR color images are fed into the proposed monocular depth estimation network to obtain the corresponding estimated depth maps. We introduce three losses to train the network: 1) reconstruction loss between the estimated LR depth and the ground-truth LR depth; 2) reconstruction loss between the downsampled estimated HR depth and the ground-truth LR depth; 3) consistency loss between the estimated LR depth and the downsampled estimated HR depth. In addition, we design a depth reconstruction network from depth to depth. Through distillation loss, features between two networks  maintain the structural consistency  in affinity space, and finally the depth reconstruction network improve the monocular depth estimation network performance.
Experimental results demonstrate that our method achieves superior performance than unsupervised and semi-supervised learning based schemes, and is competitive or even better compared to supervised ones.
\end{abstract}


\IEEEpeerreviewmaketitle

\section{Introduction}

\IEEEPARstart{3}D environment perception plays a critical role in many real world applications, such as robotics and autonomous driving, where one essential component is depth estimation. Among various depth estimation strategies, one interesting problem is to infer 3D geometric information from a 2D scene image, which is called monocular depth estimation. Due to the nature that a single 2D image may be produced from an infinite number of distinct 3D scenes, monocular depth estimation is ill-posed and thus challenging. It is an active research topic in computer vision, and draws many attentions from both academia and industry.

To solve this ill-posed problem, in the literature, many monocular depth estimation methods have been reported.
Specially, thanks to the rapid development of convolutional neural networks (CNN), learning-based approach for monocular depth estimation becomes popular. The learning strategy is ranging from supervised, semi-supervised, to unsupervised learning. The supervised manner assumes the availability
of a large training set of RGB-depth pair images, according to which depth estimation can be regarded as a pixel-level regression problem. The method proposed by Eigen et al.~\cite{eigen2014depth} is the first one that attempts to solve the monocular depth estimation problem by CNNs in an end-to-end manner, where the ground-truth depth maps serve as the supervised information. After that, dozens of supervised learning based methods are proposed with different networks architectures and loss functions~\cite{liu2018planenet,hao2018detail,lee2019monocular,zhang2019pattern,fu2018deep}, and achieve promising estimation performance. However, the success of such methods relies on a large amount of per-pixel ground-truth depth data, which are often prohibitively expensive to acquire. A depth sensor that works in an  active  sensing  manner, such as LiDAR, is commonly used to acquire depth maps, but they are usually with lower resolution and much sparser than color images, leading to much trouble on obtaining ground-truth supervision. This is why all existing public datasets  provide only a small amount of depth maps with limited scenes.

Instead of using massive ground-truth depth maps to train the prediction model, a large research effort is focused on unsupervised manner, which leverages the geometric constraints between frames or views as the supervisory information. For instance, Zhou et al.~\cite{zhou2017unsupervised} proposed a depth prediction network based on projection function between neighboring frames of monocular image sequences, and define the reconstruction loss using view synthesis error as a metric. Godard et al.~\cite{godard2017unsupervised} proposed a training loss that enforces consistency between the disparities produced relative to both the left and right
images.  However, the performance of unsupervised methods is far from the supervised counterpart. Since there is no ground truth used during training,  unsupervised methods suffer from some limitations, such as scale ambiguity and scale inconsistency. To achieve a trade-off between accuracy and independence of the ground truth, some works turn to consider the semi-supervised approach. For instance, Kuznietsov et al.~\cite{kuznietsov2017semi} proposed to use sparse ground-truth depth for supervised learning. Cho et al.~\cite{cho2019large} proposed to first train the stereo teacher network fully utilizing the binocular perception of 3D geometry, and then use depth predictions of the teacher network for supervising the student network for monocular depth inference.

In this paper, we propose a novel and effective method for monocular depth estimation, inspired by the observation of the resolution-mismatch effect of consumer-grade RGB-D cameras, such as Microsoft Kinect. These cameras are equipped with two sensors, which can capture RGB image and depth map meanwhile. Among them, the RGB sensor, such as CCD/CMOS, works in  a  passive  sensing  manner,  which records light reflected from the surface of objects. Thus, the acquired RGB images exhibit  rich structure/texture information and are with high-resolution (HR). In contrast,  the  depth sensor, such as Time-of-Flight,   works  in an  active  sensing  manner, which records an indirect measurement of the time it takes the light to travel from the camera to the scene and back. Each pixel of depth map encodes the distance to the corresponding point in the scene. The acquired depth maps thus exhibit piecewise smooth property and are with low-resolution (LR). The  difference  in  imaging  mechanism leads to resolution gap between RGB image and depth map. Instead of using pairs of HR RGB-depth images for training, as done in existing supervised methods, we propose a framework which trains the depth prediction model relying on pairs of HR RGB image and LR depth map. According to this setting, two deep neural networks are tailored, the one is monocular depth estimation network and the other is depth reconstruction network for distillation. HR and LR color images are separately input to the monocular depth estimation network to predict HR and LR depth maps. The estimation network is trained with shared weights to resolve the gap between different resolutions. We design three loss functions between predicted depth maps and  ground-truth which is LR depth map. We additionally use LR depth map to pretrain a depth reconstruction network for distillation to improve the performance of the estimation network. The input and the ground-truth are LR depth map. The mission of depth reconstruction network is to extract feature from depth map and rebuild back to depth map. We project the features of the two networks into affinity space, and maintain the structural consistency between them by distillation loss. In the test phase, we only need to input HR color image to estimate HR depth map. Since our framework does not need to access the ground-truth HR depth data, our method belongs to a weakly-supervised manner. To the best of our knowledge, this is the first work in the literature that take such an approach for monocular depth estimation. Experimental results demonstrate that our method achieves superior performance than unsupervised and semi-supervised learning based schemes, and is competitive or even better compared to supervised ones.

We highlight the main contributions of this work as follows:
\begin{itemize}
    \item We provide experimental analysis about the resolution-mismatch effect, and demonstrate that this would incur severe domain adaptation issues for supervised monocular depth estimation methods. 
    \item We propose a novel and effective weakly-supervised framework. For the first part, We design a monocular depth estimation network trained with resolution-mismatched data, and  three loss functions.  
    \item We design a depth reconstruction   network from depth-to-depth auto-encoder for the second part of weakly-supervised framework. Through distillation, the two networks can maintain the structural  between color images and its associated depth  maps  in  affinity  space. Experimental results demonstrate that the proposed weakly-supervised framework achieves promising performance compared with the state-of-the-art methods.
\end{itemize}

\section{Related Work}
\subsection{Depth Estimation}
Depth Estimation is essential for understanding the 3D structure scenes from the 2D images, which has attracted considerable attention in the last decade. The origin work mostly relied on the stereo vision~\cite{sinz2004learning}, which uses image pairs to reconstruct depth images. With the development of deep learning,~\cite{memisevic2011stereopsis} proposed a method based on deep learning to handle the stereo vision. Hoiem et al.~\cite{hoiem2005automatic} introduce photo pop-up, a fully automatic method for creating a basic 3D model from a single photograph. In a seminal work~\cite{saxena2005learning}, a supervised learning model was used to estimate depth from a single image, which caused many repercussions. Their work has been later extended to 3D scene reconstruction~\cite{saxena2008make3d}. Science then, many methods based on deep learning have been proposed to apply in monocular depth estimation. Liu et  al.~\cite{liu2010single} combine the task of semantic segmentation with depth estimation. Ladicky et al.~\cite{ladicky2014pulling} instead jointly predict labels and depth in a classification approach. 

Recently, various convolutional Neural Networks based techniques for monocular depth estimation have been proposed. Eigen et al.~\cite{eigen2014depth} used  AlexNet~\cite{krizhevsky2012imagenet} structure for global depth prediction, and added a local depth network to fine tune the global depth. Laina et al.~\cite{laina2016deeper} proposed a depth prediction method with fully convolutional residual networks. The approach of Li et al.~\cite{li2015depth} combined deep learning features on image patches with hierarchical CRFS defined on superpixel segmentation of the image. Their framework about surface normal inspired some later work. Liu et  al.~\cite{liu2018planenet} proposed a DNN for piece-wise planar depth map reconstruction from a single RGB image. Fu et al.~\cite{fu2018deep} proposed a deep ordinal regression network (DORN), which transform the depth regression task into a  classification problem. ~\cite{lee2019monocular} used relative depth maps for monocular depth estimation by a novel algorithm. Zhang et  al.~\cite{zhang2019pattern} jointly predicted depth, surface normal, and semantic segmentation through a novel Pattern-Affinitive Propagation Framework. Yin et al.~\cite{yin2019enforcing}showed the importance of the high-order 3D geometric constraints for depth prediction. By designing a loss term that enforces one simple type of geometric constraints,  namely,  virtual normal directions determined by randomly sampled three points in the reconstructed 3D space, they considerably improved the depth prediction accuracy. Huynh~\cite{huynh2020guiding} introduced guiding depth estimation to favor planar structures that are ubiquitous, especially in indoor environments. This was achieved by incorporating a non-local coplanarity constraint to the network with a novel attention mechanism called depth-attention volume. Bhat et al.~\cite{bhat2021adabins} designed atransformer-based architecture block that divides the depth range into bins whose center value is estimated adaptively per images. Song et al.~\cite{song2021monocular} proposed a simple but effective scheme by incorporating the Laplacian pyramid into the decoder architecture, which fully utilize underlying properties of well-encoded features for monocular depth estimation.

Unlike supervised methods, there almost no recent methods attempt to learn depth map prediction in a weakly supervised way, and few semi-supervised methods. Xie et al.~\cite{xie2016deep3d} used an image alignment loss in a convolutional encoder-decoder architecture but additionally enforce left-right consistency of the predicted disparities in the stereo pair. Kuznietsov~\cite{kuznietsov2017semi} used sparse ground-truth depth for supervised learning, and enforce the deep network to produce photo consistent dense depth maps in a stereo setup using a direct image alignment loss. \cite{tosi2019learning} represented a novel deep architecture designed to infer depth from a single input image by synthesizing features from a different point of view, horizontally aligned with the input image, performing stereo matching between the two cues. Amiri et al.~\cite{amiri2019semi} proposed a loss function that explicitly exploits left-right consistency in a stereo reconstruction, and describe the correct use of ground-truth depth derived from LiDAR that can significantly reduce prediction error.  Cho et  al.~\cite{cho2019large} proposed a student-teacher strategy in which a shallow student network is trained with the auxiliary information obtained from a deeper and accurate teacher network.

Unsupervised learning-based methods have emerged as an alternative to reduce the requirement of ground truth data~\cite{zhou2017unsupervised,godard2017unsupervised}. The unsupervised method uses the stereo image reconstruction principle to learn depth prediction. Therefore, the depth image of the whole image can be estimated. However, the accuracy of unsupervised depth estimation is limited by the accuracy of stereo reconstruction. For instance, Godard et al.~\cite{godard2017unsupervised} offerd an approach to learn single-view depth estimation using rectified stereo input during training.The disparity matching problem in a rectified stereo pair requires only a one-dimensional search. Dai et al.~\cite{dai2019mvs2} proposed the first unsupervised MVS network with a symmetric unsupervised network that enforces cross-view consistency of multi-view depth maps during both training and testing. Following ~\cite{godard2017unsupervised,ummenhofer2017demon}, Zhou et al.~\cite{zhou2017unsupervised} generalized this to self-supervised training in the purely monocular setting, where a depth and pose network are simultaneously learned from unlabeled monocular videos.

\subsection{Knowledge Distillation}
Knowledge distillation~\cite{hinton2015distilling} is to transfer the knowledge from the data learned by the teacher network to the student network, so that the accuracy of student is close to the teacher. In later works, the main direction to improve the knowledge distillation is to transfer more knowledge from the teacher to the students, such as intermediate representation~\cite{romero2014fitnets} and relation among instances~\cite{peng2019correlation}. Similarity-preserving knowledge distillation~\cite{tung2019similarity} required the student to mimic the pairwise similarity map between the instance features of the teacher. Yim~\cite{yim2017gift} defined the distills knowledge to be transferred in terms of flow between layers, which  is calculated by computing the inner product between features from two layers.
In the aspect of depth estimation task, \cite{guo2018learning} proposed to use the stereo matching network as a proxy to learn depth from synthetic data and use predicted stereo disparity maps for supervising the monocular depth estimation network. Cross-domain synthetic data can be fully utilized in this novel framework.  ~\cite{sun2021learning} designed a cross-task distillation scheme that encourages depth super resolution and depth estimation networks to learn from each other in a teacher-student role-exchanging fashion.

\begin{figure*}[ht!]
\centering
\includegraphics[scale=0.54]{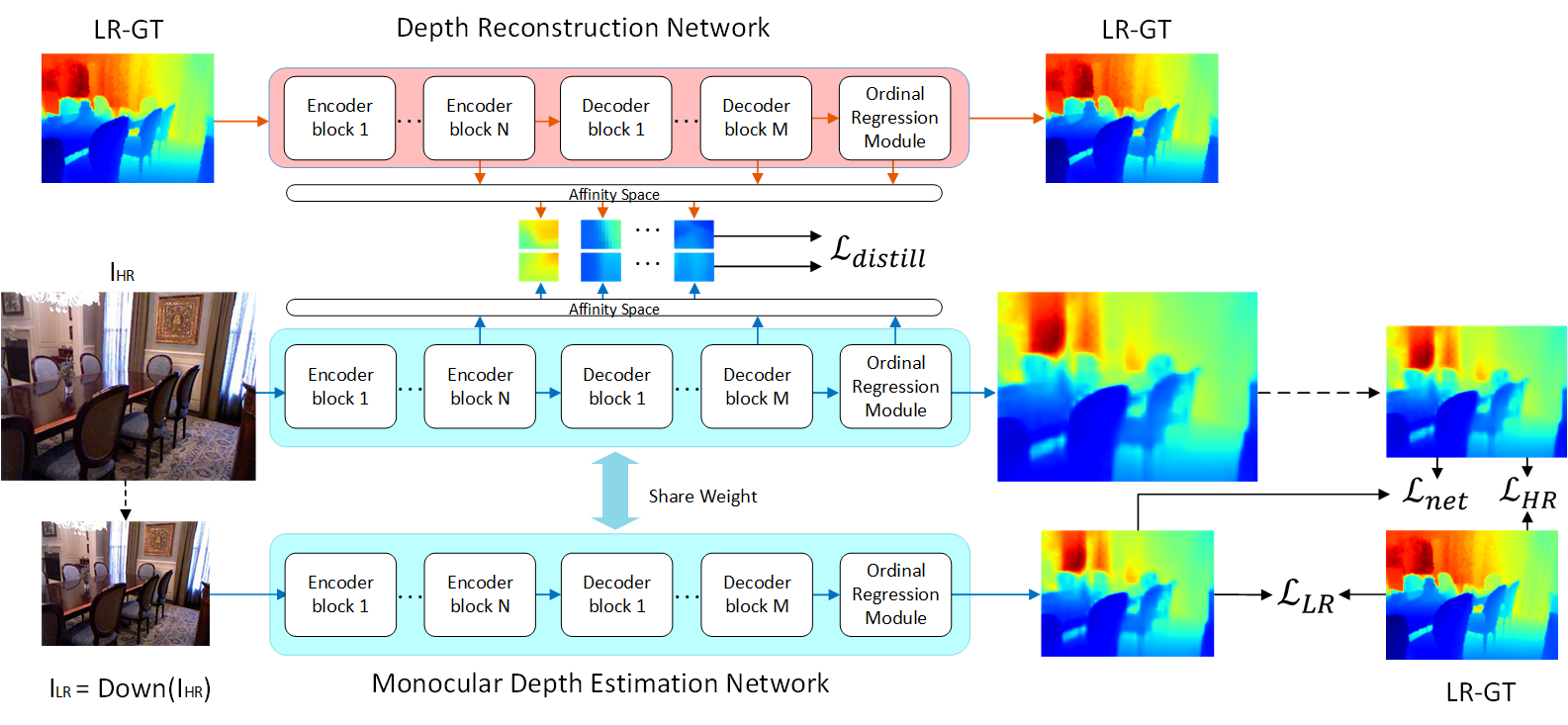}

\caption{ Illustration of the framework architecture. The size of HR images is set as \emph{W}$\times$\emph{H}, and the size of LR images is set as $\mu$\emph{W}$\times$$\mu$\emph{H}. In our weakly supervised framework, there is no HR depth images as ground-truth. Instead, ground-truth is LR depth images. $\mu$ is an adjustable down sampling multiple. In our experiments, $\mu$ is set to 0.5 and 0.25.}
\label{fig:Architecture}
\end{figure*}

\section{Proposed Approach}
In this section, we first provide the experimental analysis about the domain adaption problem incurred by resolution-mismatch between training and testing. We then elaborate the proposed weakly-supervised scheme based on resolution-mismatched data.
\subsection{Domain Adaption Incurred by Resolution-mismatch}
Even though supervised learning based monocular depth estimation methods achieve promising performance, they all have the requirement that the input RGB images and the corresponding ground-truth depth maps should share the same resolution. However, due to the difference in imaging mechanism, the RGB images are easily to capture with high resolution while depth maps are with lower-resolution. If we want to obtain depth maps with the same resolution as RGB images, a more precise and expensive depth sensor is needed. An alternative approach to remedy this problem is to downsample RGB images such that RGB-D resolution is matched. However, we demonstrate by experimental analysis that this would incur severe domain adaptation issues, if the resolution between training data and testing data is mismatched.

Specifically, we choose three representative state-of-the-art supervised monocular depth estimation methods as examples, including DPDE \cite{hao2018detail}, EGCV \cite{yin2019enforcing} and DORN ~\cite{fu2018deep}. We conduct experiments on two scenarios: 
\begin{enumerate}
\item perform training on HR RGB-D pairs and testing on HR RGB images; 
\item  perform training on LR RGB-D pairs but testing on HR RGB images. 
\end{enumerate}
As shown in Table~\ref{tab:Gap}, scenario 1) performs much better than scenario 2), \emph{i.e.}, training on HR RGB-D pairs achieves much better depth estimation performance than training on LR RGB-D pairs when testing on HR images. These results demonstrate that there is a non-negligible domain gap between HR RGB-D data and LR RGB-D data.

\begin{table}[!t]
\begin{center}
\caption{\label{tab:Gap}{Domain adaption incurred by resolution-mismatch in training and testing.} }
\begin{tabular}{c||c|c||c|c|c}
  \hline
\multirow{2}{*}{Method}&
\multirow{2}{*}{Train}&
\multirow{2}{*}{Test}&
\multicolumn{2}{c|}{Error}&
\multicolumn{1}{c}{Accuracy}\\
\cline{4-6} 
    & & &REL&RMSE& $\delta<1.25$   \\ 
\hline
\multirow{2}{*}{DPDE~\cite{hao2018detail}}&
HR&
\multirow{2}{*}{HR}&
0.127 &0.555 &	0.841 \\
\cline{2-2}
\cline{4-6}
&LR&&0.173	&0.772	&0.687\\
\hline
\multirow{2}{*}{EGCV~\cite{yin2019enforcing}}&
HR&
\multirow{2}{*}{HR}&
0.108 &	0.416 &	0.875 \\
\cline{2-2}
\cline{4-6}
&LR &&0.242&	0.936&	0.641	\\
\hline
\multirow{2}{*}{DORN~\cite{fu2018deep}}&
HR&
\multirow{2}{*}{HR}&
0.115 &	0.509 &	0.828\\
\cline{2-2}
\cline{4-6}
&LR&&0.414&	1.463&	0.079\\
\hline
\end{tabular}
\end{center}
\end{table}

\begin{figure}[ht!]
\centering
\includegraphics[scale=0.52]{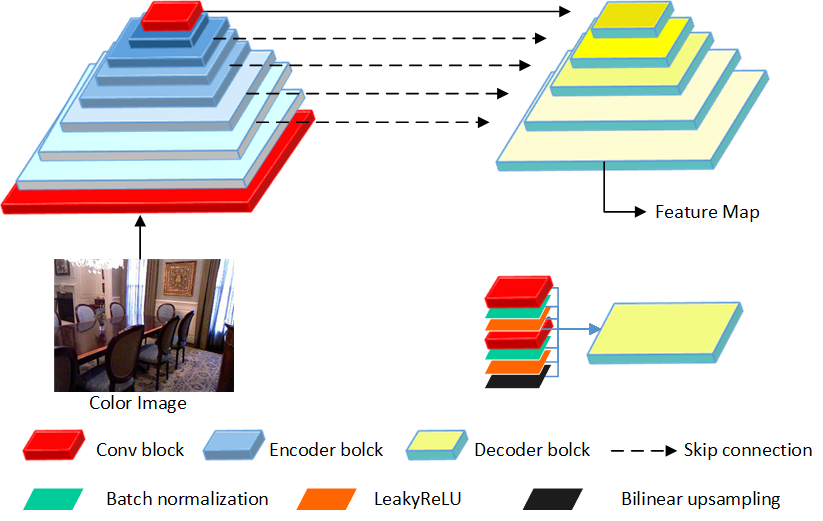}

\caption{ Illustration of the Encoder and Decoder Architecture.}
\label{fig:encoder_decoder}
\end{figure}

\subsection{Overview of Weekly-Supervised Framework}

Instead of using pairs of HR RGB-D images for training, we propose a \textit{weakly-supervised} framework that trains the depth prediction model relying on pairs of HR RGB images and LR depth maps. These coupled data can be easily obtained from consumer-grade RGB-D camera. As illustrated in Fig.\ref{fig:Architecture}, the proposed framework consists of two parts: resolution-mismatched monocular depth estimation network and  depth reconstruction network for distillation.
\subsubsection{Resolution-Mismatched Monocular Depth Estimation Network}
The proposed monocular depth estimation network(MDEN) performs training in an end-to-end manner, where the input is the HR RGB image ${\textbf{I}}_{\text{HR}}$ and the output is supervised by the LR depth map ${\textbf{D}}_{\text{LR}}$ instead of the ground-truth HR depth map ${\textbf{D}}_{\text{HR}}$. Firstly, we downsample the input color image ${\textbf{I}}_{\text{HR}}$ to obtain the LR image ${\textbf{I}}_{\text{LR}}$ with the same resolution as ${\textbf{D}}_{\text{LR}}$:
\begin{equation}
{\textbf{I}}_{\text{LR}}  = \mathrm{Down}({\textbf{I}}_{\text{HR}})
\end{equation}
where $\mathrm{Down}(\cdot)$ denotes the downsampling operator. Then, we feed both ${\textbf{I}}_{\text{HR}}$ and ${\textbf{I}}_{\text{LR}}$ into the proposed monocular depth estimation network (MDEN) to obtain the corresponding estimated depth maps $\widehat {\textbf{D}}_{\text{HR}}$ and $\widehat {\textbf{D}}_{\text{LR}}$:
\begin{align}
\widehat {\textbf{D}}_{\text{HR}}  = \mathrm{MDEN}({\textbf{I}}_{\text{HR}},\boldsymbol{\theta})\\
\widehat {\textbf{D}}_{\text{LR}}  = \mathrm{MDEN}({\textbf{I}}_{\text{LR}},\boldsymbol{\theta})
\end{align}
where $\boldsymbol{\theta}$ denote the network parameters.
We then define the following loss functions:
\begin{equation}
\begin{aligned}
\mathcal{L}_{MDEN}=\alpha \mathcal{L}_{LR}+\beta\mathcal{L}_{HR}+\gamma \mathcal{L}_{net}
\label{MDEN_loss}
\end{aligned}
\end{equation}
where the first term is the \textit{LR reconstruction loss} $\mathcal{L}_{LR}$, which denotes the error between the predicted depth ${\widehat{\textbf{D}}_{\text{LR}}}$ predicted from LR color image and the ground-truth LR depth ${{\textbf{D}}_{\text{LR}}}$; the second term is the \textit{HR reconstruction loss} $\mathcal{L}_{HR}$, which denotes the error between the downsampled version of the depth ${\widehat{\textbf{D}}_{\text{HR}}}$ predicted from HR color image and the ground-truth LR depth ${{\textbf{D}}_{\text{LR}}}$; the third term is the \textit{network consistency loss} $\mathcal{L}_{net}$, which denotes the error between the downsampled version of the HR depth predicted from HR color image and the depth predicted from LR color image.
$\mathcal{L}_{LR}$ and $\mathcal{L}_{HR}$ are defined as reconstruction loss, the details of which will be introduced in next subsection; $\mathcal{L}_{net}$ is defined as L2 loss:
\begin{equation}
\begin{aligned}
\mathcal{L}_{net}=  \|\mathrm{Down}(\widehat {\textbf{D}}_{\text{HR}})- \widehat {\textbf{D}}_{\text{LR}}\|_2^2 
\label{L2}
\end{aligned}
\end{equation}

\subsubsection{Depth reconstruction network for distillation}
Beyond the proposed resolution-mismatched monocular depth estimation network, we propose a cross-task distillation scheme.  The knowledge distillation network  take full advantage of LR depth supervision, in order to guide our LR depth estimation. We design a depth reconstruction network(DRN) from LR depth-to-depth auto-encoder as a teacher of MDEN, and its input and output are LR ground truth ${{\textbf{D}}_{\text{LR}}}$ and LR reconstructed depth map ${{\textbf{D}}_{\text{re}}}$ :
\begin{equation}
{\textbf{D}}_{\text{re}} = \mathrm{DRN}({\textbf{D}}_{\text{LR}})
\end{equation}
The goal of DRN is to obtain the features of the depth map and restore the features to a depth map close to the ground truth. To guide the prediction of the depth map better, we use a loss function $\mathcal{L}_{distill}$, witch is able to strengthen the feature structural consistency between the MDEN and DRN in the  affinity space. The architecture of DRN and $\mathcal{L}_{distll}$ will be introduced in detail in next section.

Finally, we use LR REN to guide LR MDEN to obtain a high-quality  depth map. Resolution-mismatched   monocular   depth   estimation network solves the gap mentioned above and output estimated HR depth maps. During training, we minimize the loss function  \begin{align}
\mathcal{L}  =\mathcal{L}_{MDEN}  + \mathcal{L}_{distill}  
\label{total_loss}
\end{align}
and  update the network parameters with back-propagation to achieve the optimized depth estimation network. In the testing stage, we only take the HR color image as the input of MDEN comparing with the state-of-the-art methods, which outputs the estimated HR depth map.

\subsection{Monocular Depth Estimation Network Architecture}
Fig.\ref{fig:Architecture} illustrates the architecture of our monocular depth estimation network (MDEN), including three main modules: an encoder extracting features from color images, an standard feature upsampling decoder, and an ordinal regression module computing multi-channel dense ordinal labels. 
Please note that the proposed MDEN is shared by both the HR and LR color images.
\subsubsection{Encoder and Decoder}

Some previous depth estimation methods~\cite{eigen2014depth,eigen2015predicting} use fully-connected layers as in a typical classification network, which are limited to a specific input size. Some previous depth estimation methods use high dilation rate in the encoder~\cite{yang2018deep}, which is convenient to control the resolution of feature map and adjust the convolution kernel's field-of-view in order to extract multi-scale information. However, it is difficult for these methods to obtain high frequency information, resulting in gridding artifacts. To alleviate the above problems, we design our encoder and decoder architecture as shown in Fig.~\ref{fig:encoder_decoder}. We use pretrained EfficientNet B5~\cite{tan2019efficientnet} as the feature extractor, and it contains 7 blocks and we remove the last global pooling layer and classifier layer. Each standard decoder block consists of bilinear interpolate and convolutional layers, batch normalization, activation function. Between encoder and decoder module, we add skip connections to form a rich feature. Features are concatenated together in the channel dimension combining global features and local features. The size of the decoder output is {$h$}$\times${$w$}$\times${$C$}, where $h$ and $w$ is the same as input color images.


\subsubsection{Ordinal Regression Module}
Inspired by DORN~\cite{fu2018deep}, we discretize depth and recast depth network learning as an ordinal regression problem, which is useful to handle color images of different resolutions. Specifically, we turn the standard regression problem into a multi-class classification problem, and ordinal regression module adopts adaptive binning structure (Adabin) ~\cite{bhat2021adabins}.

\begin{figure}[t!]
\centering
\includegraphics[scale=0.52]{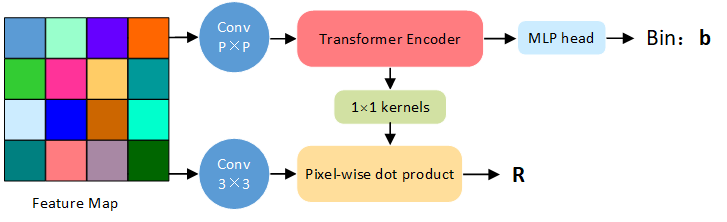}

\caption{ Illustration of the Ordinal Regression Module Architecture.}
\label{fig:Architecture_regresion_module}
\end{figure}
The structure of Ordinal Regression Module is shown in Fig.~\ref{fig:Architecture_regresion_module}, the decoded features are first pass through a convolutional layer with kernel size $p$ $\times$ $p$, stride $p$, and are reshaped into patch embeddings. We add learned positional encodings to the patch embeddings before feeding them to the transformer encoder, which is four layers Vision Transformer ViT~\cite{dosovitskiy2020image}. We use an multi-layer perceptron head over the first output embedding to ouput an N-dimensional vector \textbf{b}, and discretize depth $D$ = ($d_{min}$,$d_{max}$) into $N$ bins. 
The transformer decoded features contain local pixel-level information, and output embeddings effectively containing more global informations. The output embeddings are used as a set of 1 × 1 convolutional kernels, and are convolved with the decoder features (following a 3 $\times$ 3 convolutional layer) to obtain the Range-Attention Maps $\mathcal{R}$. $\mathcal{R}$ are passed through a 1 $\times$ 1 convolutional layer and Softmax activation to obtain $N$-channels. We interpret the N Softmax score $p_k$ ($k$= 1,..., $N$) at each pixel as probabilities over $N$ depth-bin $\mathcal{} \textbf{b} \:=\{b_{1},b_{2}, ...,(b_{N})\}$. Finally, the final depth value $\widehat{d}$ is calculated from the linear combination as follows:
\begin{equation}
\begin{aligned}
\widehat{d}=\sum\limits_{k=1}^{N}b_{i}p_{k}
\end{aligned}
\end{equation}

\subsubsection{Loss function}
In the proposed MDEN, the reconstruction losses $\mathcal{L}_{HR/LR}$ are specified in terms of Scale-Invariant loss $\mathcal{L}_{pixel}$~\cite{eigen2014depth} and bi-directional Chamfer Loss $\mathcal{L}_{bins}$~\cite{fan2017point}.
We define the reconstruction losses as:
\begin{align}
\mathcal{L}_{LR/HR}  =\mathcal{L}_{pixel} + \lambda\mathcal{L}_{bins}
\end{align}

$\mathcal{L}_{pixel}$ is a scaled version of the Scale-Invariant loss:
\begin{align}
\mathcal{L}_{pixel} =a \sqrt{\frac{1}{T}\sum_i g_{i}^{2} - \frac{b}{T^2}(\sum_i g_i)^2}
\end{align}
where $g_i = \log \hat{d_i} - \log d_i$, the ground truth depth $d_i$, predicted depth $\widehat{d}$ and $T$ denotes the number of pixels having valid ground truth values. We use $b = 0.85$, $a = 10$ for all our experiments. 

We use the bi-directional Chamfer loss~\cite{fan2017point}  as a regularizer:
\begin{align}
\mathcal{L}_{bins}=\sum\limits_{x\in X}\min_{y\in \textbf{b}}\|x-y\|^{2} + \sum\limits_{y \in \textbf{b}}\min_{x\in X}\|x-y\|^{2}   
\end{align}
The $\mathcal{L}_{bins}$ can encourage the distribution of bin centers to follow the distribution of depth values in the ground truth. We would like to decrease the error between predicted depth maps and ground truth by supervising the bin centers to be close to the actual depth values. We denote the set of bin centers as \textbf{b}, the set of all depth values in the ground truth image as $X$.


$\mathcal{L}_{LR}$ computes the  error between the predicted depth ${\widehat{\textbf{D}}_{\text{LR}}}$ from LR color image and the ground-truth LR depth ${{\textbf{D}}_{\text{LR}}}$. $\mathcal{L}_{HR}$ computes the error between the downsampled version of the predicted depth ${\widehat{\textbf{D}}_{\text{HR}}}$ from HR color image and the ground-truth LR depth ${{\textbf{D}}_{\text{LR}}}$. We use $\lambda = 0.1$ for all our experiments.

\begin{table*}[t!]
\begin{center}
\caption{\label{tab:result-kitti} Quantitative Evaluations On The KITTI Dataset~\cite{geiger2013vision}. Our weakly-supervised method achieves better objective performance than unsupervised and semi-supervised methods, and achieves competitive or even better performance compared with fully-supervised ones.} 
\begin{tabular}{c||c||c||c|c|c|c||c|c|c}
  \hline
\multirow{2}{*}{Method}&
\multirow{2}{*}{Reference}&

\multirow{2}{*}{Type}&
\multicolumn{4}{c||}{Error (lower is better)}&
\multicolumn{3}{c}{Accuracy (higher is better)}\\
\cline{4-10}
     &&&REL&Sq Rel&RMSE&RMSE log&$\delta<1.25$&$\delta<1.25^{2}$&$\delta<1.25^{3}$\\
\hline
Sexena et  al.~\cite{saxena2005learning}&NIPS'2005&fully-supervised & 0.280 & 3.012 & 8.734 & 0.361 & 0.601 & 0.820 & 0.926\\
\hline
Eigen et  al.~\cite{eigen2014depth}&NIPS’2014&fully-supervised & 0.203 & 1.548 & 6.307 & 0.282 & 0.702 & 0.898 & 0.967\\
\hline
Liu et  al.~\cite{liu2015learning}&TPAMI'2015&fully-supervised & 0.201 & 1.584 & 6.471 & 0.273 & 0.680 & 0.898 & 0.967\\
\hline
Gan et  al.~\cite{gan2018monocular}&ECCV'2018 &fully-supervised& 0.098& 0.666 & 3.933 &0.173 & 0.890 & 0.964 & 0.985\\
\hline
Fu et  al.~\cite{fu2018deep} &CVPR'2018&fully-supervised &0.072& 0.307 &2.727& 0.120&0.932& 0.984 &0.994\\
\hline
Yin et  al.~\cite{yin2019enforcing}&ICCV'2019&fully-supervised & 0.072 &-& 3.258 & 0.117 & 0.938 & 0.990 & 0.998\\
\hline
Ranft et  al.~\cite{Ranftl2020}& TPAMI'2020&fully-supervised& 0.062&-& 2.573& 0.092& 0.959 & 0.995 & 0.999\\
\hline
Song et  al.~\cite{song2021monocular} & TCSVT'2021&fully-supervised & 0.059& 0.201&	2.397&	0.090 & 0.965 & 0.995 &	0.999 \\
\hline
Bhat et  al.~\cite{bhat2021adabins} &CVPR'2021&fully-supervised  & \textbf{0.058} & \textbf{0.190} & \textbf{2.360} & \textbf{0.088} & \textbf{0.964} & \textbf{0.995} & \textbf{0.999}\\
\hline
\hline
Kuznietsov et  al.~\cite{kuznietsov2017semi}& CVPR'2017& semi-supervised & 0.113 & 0.741 & 4.621 & 0.189 &0.862 & 0.960 & 0.986 \\
\hline
Tosi et  al.~\cite{tosi2019learning}& CVPR'2019&semi-supervised&0.096& 0.673& 4.351& 0.184 &0.890& 0.961& 0.981\\
\hline
Cho et  al.~\cite{cho2019large}& TIP'2019 &semi-supervised & 0.095 & 0.613& 4.129 &0.175 & 0.884& 0.964 &0.986\\
\hline
Amiri et  al.~\cite{amiri2019semi}& ROBIO'2019 &semi-supervised&\textbf{0.078} &\textbf{0.417}& \textbf{3.464}& \textbf{0.126}& {0.923}& \textbf{0.984}& \textbf{0.995}\\
\hline
\hline
Godard et  al.~\cite{godard2017unsupervised}& CVPR'2017 & unsupervised &0.114 &0.898& 4.935& 0.206& 0.861& 0.949 &0.976 \\
\hline
Monodepth2~\cite{godard2019digging} & ICCV'2019 & unsupervised & 0.115 & 0.882 & 4.701 & 0.190 & 0.879 & 0.961 & 0.982\\
\hline
Guizilini et  al.~\cite{guizilini2020semantically} & ICLR'2020 & unsupervised & 0.100 & 0.761 & 4.270 & 0.175 & 0.902 & 0.965 & 0.982\\
\hline
Packnet-SFM ~\cite{guizilini20203d} & CVPR'2020 & unsupervised & 0.107 & 0.802 & 4.538 & 0.186 & 0.889 & 0.962 & 0.981\\
\hline
Hur et  al.~\cite{hur2020self} & CVPR'2020 & unsupervised &0.106& 0.888& 4.853 &0.175 &0.879 &0.965 &0.987\\
\hline
Shu et  al.~\cite{shu2020feature} & ECCV'2020 & unsupervised &0.104 & 0.729 & 4.481 & 0.179 & 0.893 & 0.965 & 0.984\\
\hline
Watson et  al.~\cite{watson2021temporal}& CVPR'2021& unsupervised & \textbf{0.087}& \textbf{0.685} & \textbf{4.142} & \textbf{0.167} & \textbf{0.920} & \textbf{0.968} & \textbf{0.983}\\
\hline
\hline
Ours &This paper&weakly-supervised & 0.071 & 0.237 & 2.592&	0.100 &	0.954 &	0.994	& 0.998\\
\hline
\end{tabular}

\end{center}
\end{table*}

\begin{table*}[t!]
\begin{center}
\caption{\label{tab:result-nyu}QUANTITATIVE EVALUATIONS ON THE NYU DEPTH V2 DATASET~\cite{silberman2012indoor} } 
\begin{tabular}{c||c||c||c|c|c||c|c|c}
  \hline
\multirow{2}{*}{Method}&
\multirow{2}{*}{Reference}&
\multirow{2}{*}{Type}&
\multicolumn{3}{c||}{Error (lower is better)}&
\multicolumn{3}{c}{Accuracy (higher is better)}\\
\cline{4-9}
     &&&REL&RMSE&$log_{10}$&$\delta<1.25$&$\delta<1.25^{2}$&$\delta<1.25^{3}$\\
\hline
Eigen  et   al.~\cite{eigen2014depth}&NIPS'2014&fully-supervised &0.158 & 0.641 & – & 0.769 & 0.950 & 0.988\\
\hline
Laina  et   al.~\cite{laina2016deeper}&3DV'2016& fully-supervised & 0.127  & 0.573 & 0.055 & 0.811 & 0.953 & 0.988 \\
\hline
Liu  et   al.~\cite{liu2018planenet}&CVPR'2018&fully-supervised &0.142 &0.514& 0.060 & 0.812& 0.957 &0.989\\
\hline
Fu  et   al.~\cite{fu2018deep}&CVPR'2018&fully-supervised &0.115 &	0.509 &	0.051 & 0.828& 	0.965 &	0.992\\
\hline
Hao  et   al.\cite{hao2018detail}&3DV'2018 &fully-supervised&0.127 &0.555 &0.053&	0.841& 	0.966 &	0.991\\
\hline
Yin  et   al.~\cite{yin2019enforcing}&ICCV'2019&fully-supervised &0.108 &	0.416 & 0.048 & 0.875 &	0.976 	&0.994\\
\hline
Lee  et   al.~\cite{lee2019monocular}&CVPR'2019&fully-supervised& 0.131 & 0.538 & - & 0.837 & 0.971 & 0.994\\
\hline
Zhang  et   al.~\cite{zhang2019pattern}&CVPR'2019&fully-supervised &0.121& 0.497 & - &0.846& 0.968& 0.994\\
\hline
Chen et al.~\cite{chen2019structure}&AAAI'2019&fully-supervised & 0.111& 0.514 & 0.048& 0.878 & 0.977 & 0.994\\
\hline
Huynh  et   al.~\cite{huynh2020guiding}&ECCV'2020 & fully-supervised& 0.108 & 0.412 & - & 0.882 & 0.980 & 0.996 \\
\hline
Song  et   al.~\cite{song2021monocular}& TCSVT'2021& fully-supervised & 0.110 & 0.393 & 0.047 & 0.885 & 0.979 & 0.995\\
\hline
Bhat  et   al.~\cite{bhat2021adabins}&CVPR'2021&fully-supervised& \textbf{0.103} & \textbf{0.364}& \textbf{0.044}& \textbf{0.903}& \textbf{0.984}& \textbf{0.997} \\
\hline
\hline
Kundu  et   al.~\cite{kundu2018adadepth} & CVPR'2018 & unsupervised & 0.136 & 0.603 & 0.057 & 0.805 & 0.948 & 0.982\\
\hline
\hline
Tian  et   al.~\cite{tian2019semi}&ICASSP'2019&semi-supervised & 0.204 & 0.704 & 0.083 & 0.686 & 0.917 & 0.977\\
\hline
Ji  et   al.~\cite{ji2019semi}&TPAMI'2019&semi-supervised& 0.183&0.704& 0.077 &0.713&0.931&0.984\\

\hline
\hline
Ours &This paper&weakly-supervised&0.129&	0.489&	0.059&	0.815	&0.959& 0.990\\
\hline
\end{tabular}

\end{center}
\end{table*}

\subsection{Depth  Reconstruction  Network}

A number of knowledge distillation techniques have been developed to benefit a naive model with some useful information from a performing model.  Some case-study approaches~\cite{guo2018learning,sun2021learning} were used to verify the effectiveness of distillation from different tasks network to MDEN. Different distillation methods need to design specific loss function to give full play to the best distillation effect. Guo et al.~\cite{guo2018learning} used the binocular depth prediction network to distill the monocular depth prediction network. They chose disparity map as the supervised information of the distillation network. Sun et al.~\cite{sun2021learning} adopted depth map super resolution network to be teacher, and they could directly calculate the feature error between the depth map super resolution network and MDEN. We design a teacher depth reconstruction network(DRN) and adopt suitable loss function $\mathcal{L}_{distill}$, which are able to play the role of a teacher as much as possible.
\subsubsection{Architecture of Depth  Reconstruction  Network}
Our goal is to allow the teacher model to provide more effective information to the student model. We design a depth reconstruction network (DRN) as teacher of MDEN. The depth reconstruction network performs rebuilding functions for depth images. As illustrated in Fig.\ref{fig:Architecture}, the architecture of DRN is almost the same as the MDEN except for the number of input channels. The input layer of DRN is one channel.  According to Eq.\ref{total_loss}, we take a LR depth map as input, and the output is a reconstructed depth map ${{\textbf{D}}_{\text{re}}}$ of the same resolution.
Since DRN and MDEN share the same network architecture of depth reconstruction, MDEN can learn the reconstruction process from DRN and recover the depth images close to the ground-truths. We find that DRN can achieve better performance with fewer training steps in the experiments.

\subsubsection{Loss function for distillation}
Since the input of the student network and the teacher network are color images and depth images respectively, their representations and resolution are different. It is not appropriate to directly calculate the difference between each pixel in the feature layers. 
In order to maintain the structural consistency between the color image and its associated depth map, we adopt affinity space distillation~\cite{sun2021learning}, which allows the student network to obtain more consistent information between pixels. In the affinity space, the correlation of pixels within each feature should be consistent between REN and MDEN. In the both networks, the size of the feature ({$F$}) is {$w$} $\times$ {$h$} $\times$ {$c$}. The reshape function $\mathbb{R}$ recasts $F$ as $\mathbb{R}$($F$) with the dimension of {$wh$ $\times$ $c$}. The affinity matrix $A$ is defined as:

\begin{equation}
\begin{aligned}
\text{$A$($F$)} = {\sigma} (\mathbb{R}(F) \otimes \mathbb{R}^{T}(F)),
\end{aligned}
\end{equation}
where $\sigma(\cdot)$ is the softmax operation, $\otimes$ is the matrix multiplication and $T$ is the transpose operator. The distillation loss of affinity space is defined as the following,
\begin{equation}
\begin{aligned}
\mathcal{L}_{distill}=\frac{1}{N}\sum\limits_{i=1}^{N}W_{i}{\|A(F_{DRN}^{i})-A(F_{MDEN}^{i})\|_1}
\end{aligned}
\end{equation}
In the distillation process, we use $W_{i}$ to adjust to the weight of distillation loss between different features. Like commonly used distillation techniques, the teacher network is trained beforehand and fixed when encouraging student network.

\begin{figure*}[ht!]

\centering
\includegraphics[height=19.125cm, width=17cm]{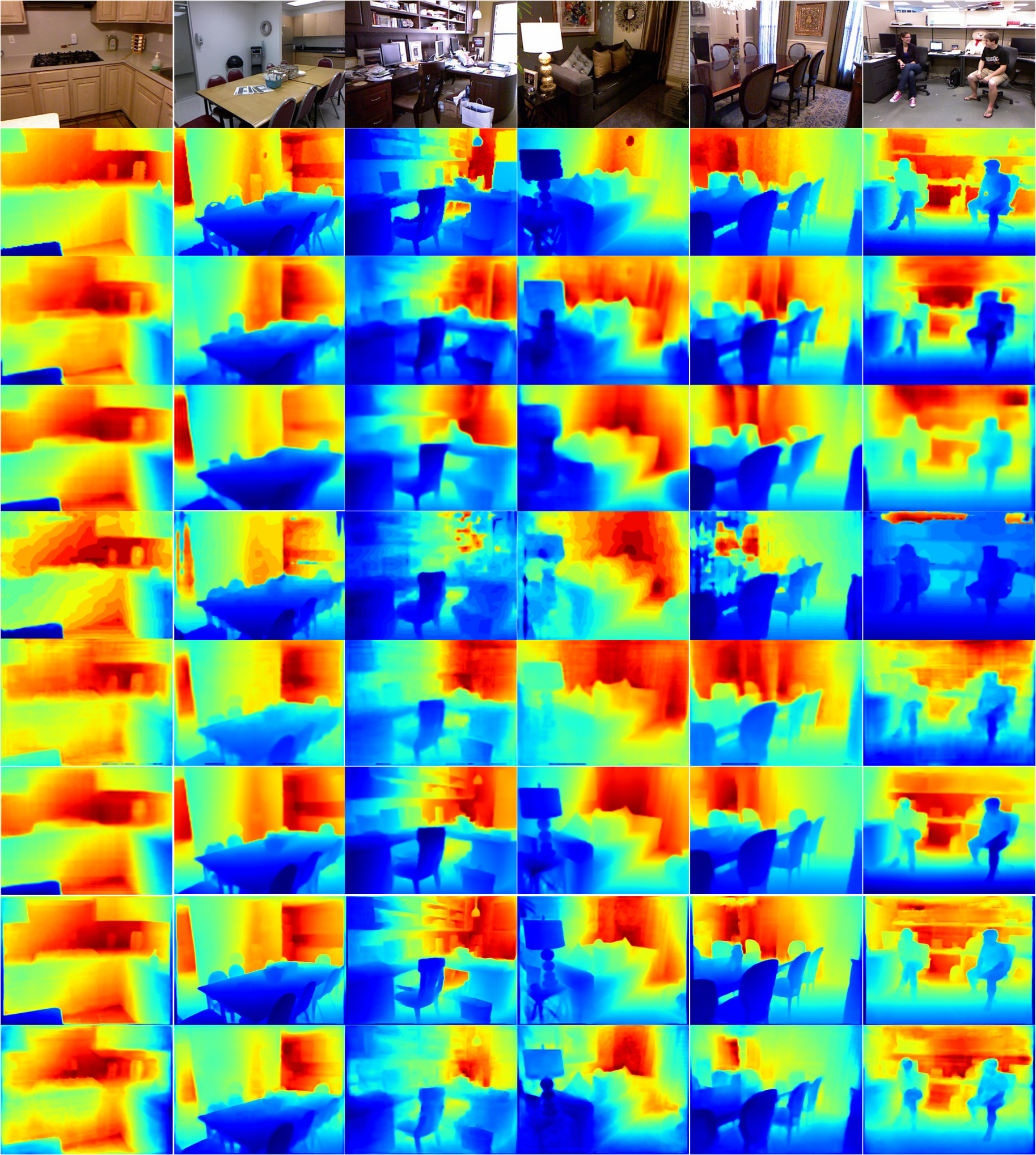}
\centering
\caption{\label{fig:images-nyu}{Subjective performance comparison with fully-supervised methods on NYU Depth-v2 dataset.} Example outputs of our method in NYU Depth-v2 dataset, as well as SOTA methods. 1st  rows: input color images. 2nd  rows: ground truth. 3rd rows: results by Eigen et al.~\cite{eigen2015predicting}. 4th rows: results by Laina et  al~\cite{laina2016deeper}. 5th rows: results by Fu et  al.~\cite{fu2018deep}. 6th  rows: results by Ramam et  al.~\cite{ramamonjisoa2019sharpnet}. 7th  rows: results by Chen et  al~\cite{chen2019structure}. 8th rows: results by Bhat et al.~\cite{bhat2021adabins}. 9th rows: results by the proposed method.  Our weakly-supervised method achieves competitive subjective results with fully-supervised ones.}
\end{figure*}

\begin{figure*}[ht!]

\centering
\includegraphics[height=20.38cm, width=16.68cm]{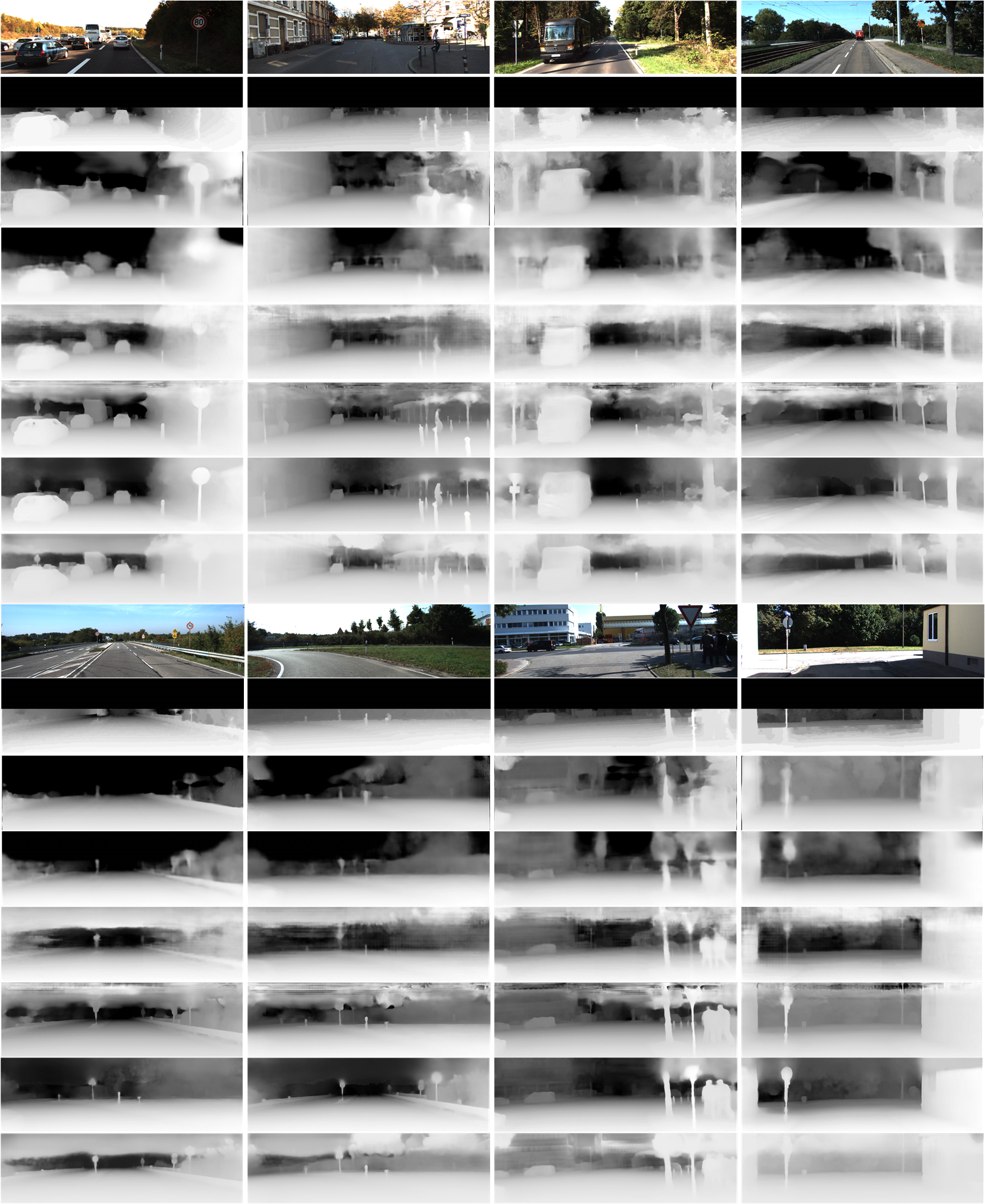}
\centering
\caption{\label{fig:images-kitti}Subjective performance comparison on KITTI dataset. Example outputs of our method in KITTI, as well as SOTA methods. 1st and 9th rows: input color images. 2nd and 10th rows: ground truth. 3rd and 11th rows: results by Godard et al.~\cite{godard2017unsupervised}. 4th and 12th rows: results by Kuznietsov et  al~\cite{kuznietsov2017semi}. 5th and 13th rows: results by Fu et  al.~\cite{fu2018deep}. 6th and 14th rows: results by Lee et  al.~\cite{lee2019big}. 7th and 15th rows: result by Song et  al~\cite{song2021monocular}. 8th and 16th rows: results by the proposed method. Our weakly-supervised method achieves competitive subjective results with fully-supervised ones.}
\end{figure*}

\section{Experiments}
In this section, we provide extensive experimental results to demonstrate the effectiveness of our proposed method. The ablation analysis is also provided for deeper understanding of our method.
\subsection{Datasets and Evaluation Metrics}
\paragraph{NYU Depth-v2~\cite{silberman2012indoor}} The NYUD-v2 dataset contains about 120K RGB and depth images obtained with a Microsoft Kinect from 464 indoor scenes. The depth maps have an upper bound of 10 meters. The image resolution is 640 $\times$ 480 pixels, in accordance with the high resolution size we set. From the entire dataset, we use 20k images for training from 249 scenes, and official test set of 654 official images from 215scenes for testing as introduced in~\cite{eigen2014depth}.

\paragraph{KITTI~\cite{geiger2013vision}} The KITTI dataset contains various road environments acquired from autonomous driving scenarios. This dataset consists of over 93K outdoor images collected form a car with stereo cameras and Lidar ground truth. In our experiment, we follow the experimental protocol proposed by Eigen et  al.~\cite{eigen2014depth}, and consider 22600 images corresponding to 32 scenes as training data and 697 images associated to other 29 scenes as test data specified by~\cite{eigen2014depth}. The maximum value of our predicted output is limited to the order of 80 meters in the test phase, as explained in the guideline of the KITTI dataset. We also adopt the central cropping scheme used in~\cite{garg2016unsupervised} for the performance evaluation.The size of HR resolution set in experiment is original 1224 $\times$ 368 pixels. 

\paragraph{Evaluation Metrics} Following previous methods~\cite{liu2018planenet,fu2018deep,hao2018detail,yin2019enforcing}, we evaluate the performance of monocular depth prediction quantitatively based on six metrics in NYU Depth-v2 dataset: 

\begin{itemize}
\item average relative error (REL): $\frac{1}{n}\sum_p^n \frac{\lvert y_p-\hat{y}_p \rvert}{y}$;
\item root mean squared error (RMS): $\sqrt{\frac{1}{n}\sum_p^n (y_p-\hat{y}_p)^2)}$;
\item average ($\log_{10}$) error: $\frac{1}{n}\sum_p^n \lvert \log_{10}(y_p)-\log_{10}(\hat{y}_p) \rvert$;
\item threshold accuracy ($\delta_i$): percentage of $y_p$ s.t. $\text{max}(\frac{y_p}{\hat{y}_p},\frac{\hat{y}_p}{y_p}) = \delta < thr$ for $thr=1.25,1.25^2,1.25^3$;
\end{itemize}
For KITTI dataset, we additionally use another two standard metrics: 
\begin{itemize}
\item Squared Relative Difference (Rel): $\frac{1}{n}\sum_p^n \frac{\|y_p-\hat{y}_p \|^2}{y}$;  
\item RMSE log: $\sqrt{\frac{1}{n}\sum_p^n \|\log y_p - \log \hat{y}_p\|^2}$.
\end{itemize}

\begin{table*}[ht]
\begin{center}
\caption{\label{tab:result-ablation}Ablation study about the roles of different losses to the final performance in NYU Depth-v2 dataset.}
\begin{tabular}{c||c|c|c||c|c|c}
  \hline
Loss used in training & REL& RMSE & $log_{10}$ & $\delta<1.25$&$\delta<1.25^{2}$&$\delta<1.25^{3}$\\

\hline
$\mathcal{L}_{LR}$&
0.180	&	0.635 & 0.075 & 0.716 &0.932 & 0.985\\
\hline
$\mathcal{L}_{LR}+\mathcal{L}_{net}$&
0.166 & 0.603&	0.072& 0.730 & 0.943 &  0.988\\
\hline
$\mathcal{L}_{LR}+\mathcal{L}_{HR}$&
0.147 & 0.536&	0.065& 0.778 & 0.960 &  0.992\\
\hline
$\mathcal{L}_{LR}+\mathcal{L}_{HR}+\mathcal{L}_{net}$&
0.135	&	0.501  &	0.061	&	0.809		&	0.961	&		0.992\\
\hline  
$\mathcal{L}_{LR}+\mathcal{L}_{HR}+\mathcal{L}_{net}+\mathcal{L}_{distill}$&
0.129 &0.489&	0.059&	0.815	&0.959 & 0.990\\ 

\hline
\end{tabular}

\end{center}
\end{table*}

\begin{table*}[h]
\begin{center}
\caption{\label{tab:result-distill_net-ablation}\noindent  The result of depth reconstruction network performances on NYU-Depth-V2 datasets. We use the LR depth images as the input and output to train the distillation network. We choose two methods, corresponding to different channel numbers of the input layer. In two ways, the depth images both can be perfectly reconstructed by the DRN}
\begin{tabular}{c|c|c|c||c|c|c}
  \hline
channel of input layer &REL  & RMSE & $log_{10}$ & $\delta<1.25$&$\delta<1.25^{2}$&$\delta<1.25^{3}$\\
\hline
three channel &0.072 & 0.247 & 0.030 & 0.969    &1 &  1\\
\hline
one channel&0.015 & 0.052  & 0.007 & 0.999 & 1 &1\\
\hline
\end{tabular}

\end{center}
\end{table*}

\begin{table*}[h]
\begin{center}

\caption{\label{tab:result-ablation2}Result of Closing Gap. We adjusted the ratio of high resolution and low resolution. And use low-resolution and high-resolution color images as input to test. }
\begin{tabular}{c||c||c|c|c||c|c|c}
  \hline
     Train&Test& REL& RMSE& $log_{10}$ & $\delta<1.25$ &$\delta<1.25^{2}$&$\delta<1.25^{3}$\\
\hline
     \multirow{2}*{LR:HR=$1:2$} & LR & 0.131& 0.498& 0.060 &	0.817&0.960 &0.990	\\ 
\cline{2-8}
     & HR & 0.129& 0.489& 0.059 &	0.815&	0.959 & 0.990 \\ 
\hline
\multirow{2}*{LR:HR=$1:4$} & LR & 0.151 & 0.519 & 0.066 & 0.805 & 0.958 & 0.990\\
\cline{2-8}
& HR & 0.149 & 0.510 & 0.065 & 0.808 & 0.951&0.989 \\
\hline
\end{tabular}
\end{center}
\end{table*}

\subsection{Implementation Details}
Our model is implemented with the Pytorch~\cite{paszke2019pytorch}. We train our model using two Nvidia-RTX 3090 with 48GB of GPU memory. For training, we use the AdamW optimizer~\cite{loshchilov2019decoupled} with weight-decay $\mathcal{}10^{-2}$. We use the 1-cycle policy~\cite{smith2018super} with initial learning rate = $\mathcal{}3.5\times 10^{-4}$. We use bilinear interpolation to be our downsample method for color images and depth images. We perform the data augmentation for the training samples by the following steps: the color and depth images are randomly flipped in the horizon, and randomly rotated from -5 to 5 degrees. The weights of MDEN are shared by joint training.  At test time, we compute the final output by taking the average of an image’s prediction and the prediction of its mirror image which is commonly used in previous work. For both datasets, we train 100 epochs with batch size 4, and each epoch spends 15 minutes. We set number of bins ($N$) to 256, which is consistent with previous work~\cite{bhat2021adabins}. We set the weight loss coefficients $\alpha$, $\beta$, $\gamma$ as 10, 1, 1. 

\subsection{Performance Comparison}
We compare our method with state-of-the-art monocular depth estimation works including unsupervised~\cite{godard2017unsupervised,guizilini20203d,hur2020self,shu2020feature,watson2021temporal,guizilini2020semantically,godard2019digging,kundu2018adadepth}, semi-supervised~\cite{tian2019semi,ji2019semi,kundu2018adadepth,kuznietsov2017semi,tosi2019learning,cho2019large} and fully-supervised ones~\cite{eigen2014depth,laina2016deeper,liu2018planenet,fu2018deep,hao2018detail,yin2019enforcing,lee2019monocular,zhang2019pattern,huynh2020guiding,bhat2021adabins,saxena2005learning,gan2018monocular,Ranftl2020,song2021monocular} on two benchmark datasets, i.e., KITTI~\cite{geiger2013vision} and NYU Depth-v2~\cite{silberman2012indoor} datasets. Since the fully-supervised methods exploit the groundtruth HR depth maps for training, in theory it should work better than other learning manners. Thus, the fully-supervised methods can be considered as the performance reference, to which we want to our performance to be as close as possible.
\subsubsection{KITTI dataset} Table~\ref{tab:result-kitti} lists the performance metrics on the KITTI dataset. Our method significantly outperforms unsupervised and semi-supervised methods. Our method can approach the sate-of-the-art, and surpass a number of fully-supervised methods. See Fig.~\ref{fig:images-kitti}, we compare the subjective effects of different methods. Following the previous works~\cite{bhat2021adabins,fu2018deep,godard2017unsupervised,song2021monocular}, the ground-truth depth map is interpolated from sparse measurements for visualization purpose. In addition to the sate-of-the-art in the seventh and fifteenth lines, our subjective renderings can exceed the methods that include full supervision.

\subsubsection{NYU Depth-v2 dataset}In Table~\ref{tab:result-nyu}, we show the quantitative evaluation on the official NYU Depth-v2 test set.  Our method is competitive with the state-of-the-arts, and can surpass some of the fully-supervised methods released in previous years. Of course, our weakly-supervised method surpasses all semi-supervised methods and unsupervised methods. We also show subjective performance comparison with other three methods on NYU Depth-v2 datset in Fig.~\ref{fig:images-nyu}. Note that all methods are fully-supervised methods except ours. Our method exhibits excellent subjective effects.

\begin{figure}[ht]
\centering
\includegraphics[scale=0.2]{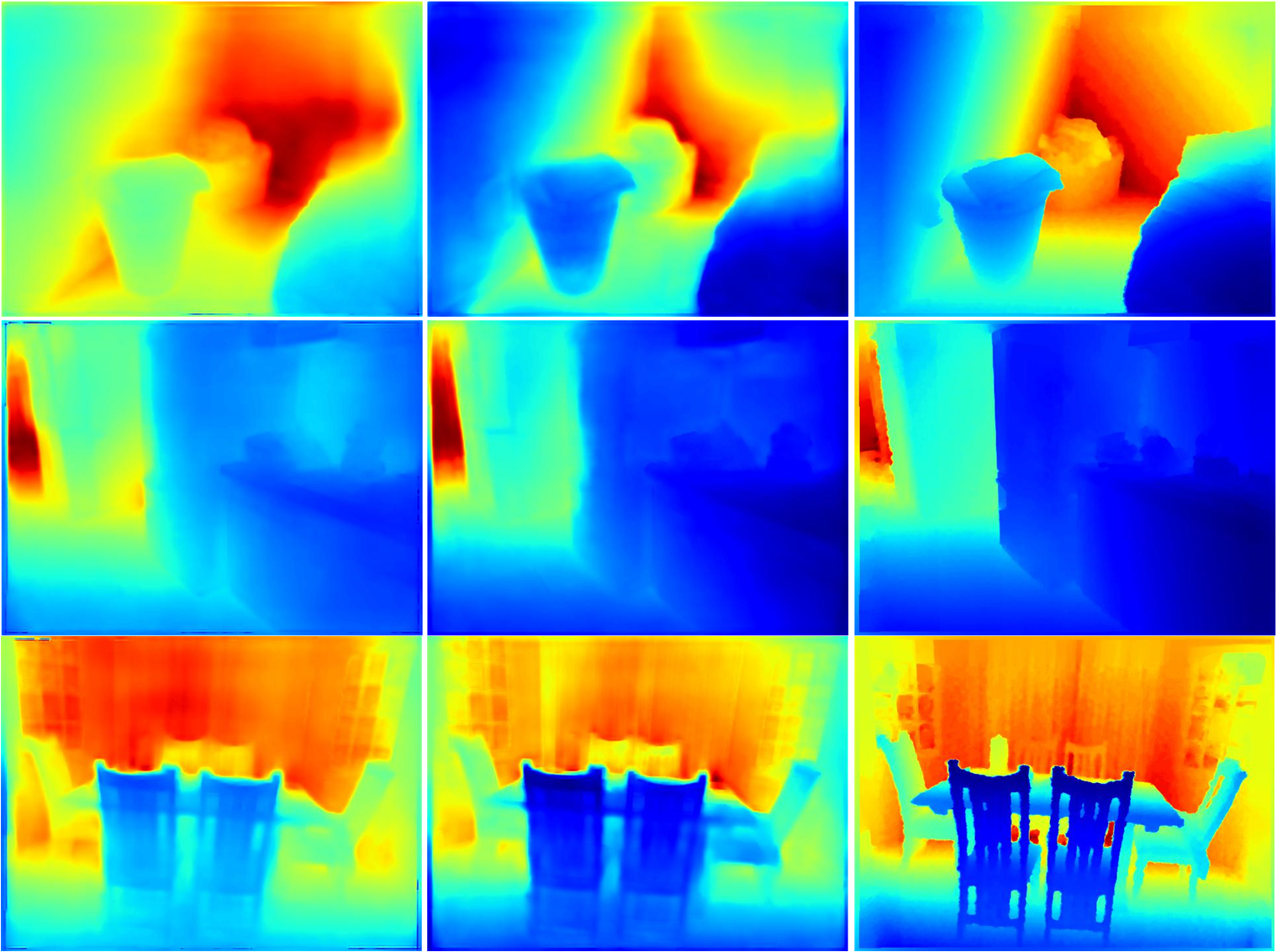}
\\
\hspace{0.02\linewidth}(a) \hspace{0.26\linewidth}(b)\hspace{0.25\linewidth}(c)\\
\caption{ {Visual comparison of depth estimation results according to different combinations of loss functions}. (a) $ \mathcal{L}_{LR}+\mathcal{L}_{net}$. (b)  $ \mathcal{L}_{LR}+\mathcal{L}_{net}+\mathcal{L}_{HR}$. (c) Ground truth. Note that the $\mathcal{L}_{HR}$ contributes to yield more reliable results.}\label{fig:HR loss}
\end{figure}

\begin{figure}[ht]
\centering
\includegraphics[scale=0.2]{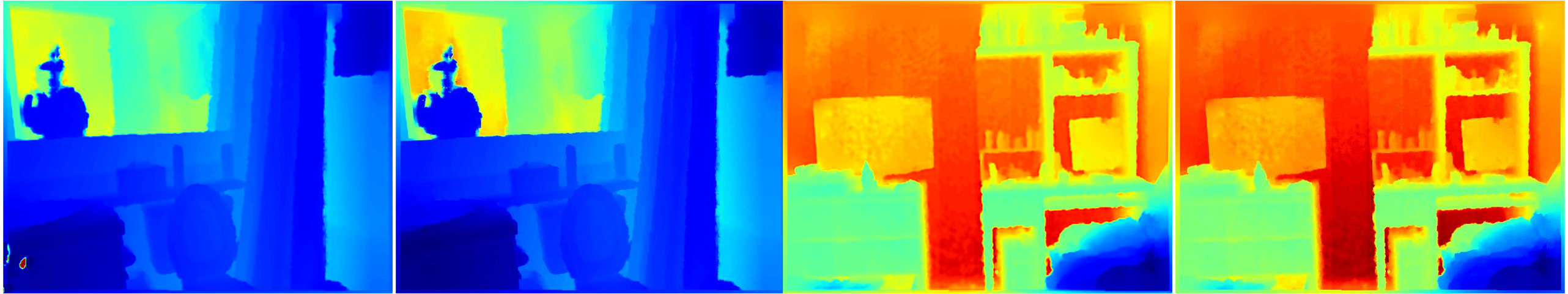}
\\
\hspace{0.02\linewidth}(a) \hspace{0.2\linewidth}(b)\hspace{0.2\linewidth}(c) \hspace{0.2\linewidth}(d)\\
\caption{ {Visual comparison between reconstructed depth images from DRN and its corresponding ground truth. } (a) and (c) are reconstructed depth images. (b) and (d) are ground truth. It can be seen that there is almost no difference between the reconstructed depth images and ground truth. }\label{fig:dis_re_result}
\end{figure}

\subsection{Ablation Study}
We conduct the following ablation studies to analyze the details of our approach. 

\paragraph{Roles of three losses to the final performance} 
In order to investigate the roles of LR reconstruction loss $\mathcal{L}_{LR}$, HR reconstruction loss $\mathcal{L}_{HR}$ and network consistency loss $\mathcal{L}_{net}$ to the final performance, we combine different loss functions to determine their impacts on the final quantitative evaluations on NYU Depth-v2 dataset shown in Table~\ref{tab:result-ablation}. It can be found that, only using $\mathcal{L}_{LR}$ achieves the worst performance. The performance is improved if $\mathcal{L}_{LR}$ and $\mathcal{L}_{net}$ is combined, and is further higher if using both $\mathcal{L}_{LR}$ and $\mathcal{L}_{HR}$. Jointly using these three losses achieves the best performance. 
On the basis of $\mathcal{L}_{LR}$ and $\mathcal{L}_{net}$, adding $\mathcal{L}_{HR}$ makes the quantitative evaluation a huge improvement,
and in NYU Depth-v2 dataset we show a comparison chart of the subjective effect of using three loss functions and no  $\mathcal{L}_{HR}$ in Fig~\ref{fig:HR loss}. We can find that the result of just using $\mathcal{L}_{LR}$ and $\mathcal{L}_{net}$ is very bad compared with the ground truth. Adding $\mathcal{L}_{HR}$ on this basis can greatly improve the subjective effect.


\paragraph{Role of the distillation network to the final performance} 
In order to allow the depth reconstruction network to better reconstruct the depth map, we design different reconstruction methods. Since we use the structure of the monocular depth prediction network as the structure of the deep reconstruction network, the only problem is that the number of channels in the input layer does not match. The depth map input to the deep reconstruction network is single-channel, while the input color image of the depth prediction network is three-channel. We try two ways to solve this problem. The first method is to keep the three channels unchanged in the input layer and copy three copies of the depth map as three-channel data into the DRN. The second method is to change the input layer of the DRN to a single channel. After training the DRN form LR depth images to LR depth images, we test the performance of two methods in NYU Depth-v2 dataset as shown in Table~\ref{tab:result-distill_net-ablation}. The first three-channel method is much worse than the second single-channel reconstruction depth. We finally adopted the second method, and we also show the subjective rendering of the reconstruction depth in Fig~\ref{fig:dis_re_result}. The reconstructed depth map is almost perfectly close to ground truth.

We further study the affect of DRN on MDEN as shown in Tab.~\ref{tab:result-ablation}. Comparing the results without DRN, adding $\mathcal{L}_{distill}$ improves the final performance. 

\paragraph{Result of closing gap} 
To further explore the effectiveness of our method, we experiment with different HR to LR ratios. As shown in Table~\ref{tab:result-ablation2}, we set the LR (depth maps) to HR (color images) ratios as $1:2$ and $1:4$ respectively. To evaluate the gap we mention above, we test the depth estimation results from both the LR and HR color images in NYU Depth-v2 dataset. The LR color images are downsampled from the HR color images. The predicted LR and HR depth maps are tested with the corresponding resolution ground truth.  Due to the effect of the network consistency loss, the quality of the LR depth maps and HR depth maps estimated with the shared monocular depth estimation network is close to each other, which demonstrates that our approach can successfully close the domain gap on resolution-mismatched data.

\section{Conclusion}
In this paper, we present a novel and effective weakly-supervised method for monocular depth estimation and it is composed of resolution-mismatched monocular depth estimation and depth reconstruction network for distillation. Our scheme trains the depth prediction model relying on pairs of HR RGB images and LR depth maps. 
For monocular depth estimation network, three loss functions including LR and HR reconstruction losses and consistency loss are tailored to obtain optimal network parameters. Furthermore, the distillation method from LR depth-to-depth  auto-encoder can maintain the structural consistency between color images and its associated depth maps in affinity space.
Experimental results on benchmark datasets constructed under various indoor and outdoor environments demonstrate that the proposed method achieves competitive performance compared with the state-of-the-arts.


%

\ifCLASSOPTIONcaptionsoff
  \newpage
\fi



%

\begin{thebibliography}{10}
\providecommand{\url}[1]{#1}
\csname url@rmstyle\endcsname
\providecommand{\newblock}{\relax}
\providecommand{\bibinfo}[2]{#2}
\providecommand\BIBentrySTDinterwordspacing{\spaceskip=0pt\relax}
\providecommand\BIBentryALTinterwordstretchfactor{4}
\providecommand\BIBentryALTinterwordspacing{\spaceskip=\fontdimen2\font plus
\BIBentryALTinterwordstretchfactor\fontdimen3\font minus
  \fontdimen4\font\relax}
\providecommand\BIBforeignlanguage[2]{{%
\expandafter\ifx\csname l@#1\endcsname\relax
\typeout{** WARNING: IEEEtran.bst: No hyphenation pattern has been}%
\typeout{** loaded for the language `#1'. Using the pattern for}%
\typeout{** the default language instead.}%
\else
\language=\csname l@#1\endcsname
\fi
#2}}

\bibitem{eigen2014depth}
D.~Eigen, C.~Puhrsch, and R.~Fergus, ``Depth map prediction from a single image
  using a multi-scale deep network,'' \emph{arXiv preprint arXiv:1406.2283},
  2014.

\bibitem{liu2018planenet}
C.~Liu, J.~Yang, D.~Ceylan, E.~Yumer, and Y.~Furukawa, ``Planenet: Piece-wise
  planar reconstruction from a single rgb image,'' in \emph{Proceedings of the
  IEEE Conference on Computer Vision and Pattern Recognition}, 2018, pp.
  2579--2588.

\bibitem{hao2018detail}
Z.~Hao, Y.~Li, S.~You, and F.~Lu, ``Detail preserving depth estimation from a
  single image using attention guided networks,'' in \emph{2018 International
  Conference on 3D Vision (3DV)}.\hskip 1em plus 0.5em minus 0.4em\relax IEEE,
  2018, pp. 304--313.

\bibitem{lee2019monocular}
J.-H. Lee and C.-S. Kim, ``Monocular depth estimation using relative depth
  maps,'' in \emph{Proceedings of the IEEE/CVF Conference on Computer Vision
  and Pattern Recognition}, 2019, pp. 9729--9738.

\bibitem{zhang2019pattern}
Z.~Zhang, Z.~Cui, C.~Xu, Y.~Yan, N.~Sebe, and J.~Yang, ``Pattern-affinitive
  propagation across depth, surface normal and semantic segmentation,'' in
  \emph{Proceedings of the IEEE/CVF Conference on Computer Vision and Pattern
  Recognition}, 2019, pp. 4106--4115.

\bibitem{fu2018deep}
H.~Fu, M.~Gong, C.~Wang, K.~Batmanghelich, and D.~Tao, ``Deep ordinal
  regression network for monocular depth estimation,'' in \emph{Proceedings of
  the IEEE conference on computer vision and pattern recognition}, 2018, pp.
  2002--2011.

\bibitem{zhou2017unsupervised}
T.~Zhou, M.~Brown, N.~Snavely, and D.~G. Lowe, ``Unsupervised learning of depth
  and ego-motion from video,'' in \emph{Proceedings of the IEEE conference on
  computer vision and pattern recognition}, 2017, pp. 1851--1858.

\bibitem{godard2017unsupervised}
C.~Godard, O.~Mac~Aodha, and G.~J. Brostow, ``Unsupervised monocular depth
  estimation with left-right consistency,'' in \emph{Proceedings of the IEEE
  conference on computer vision and pattern recognition}, 2017, pp. 270--279.

\bibitem{kuznietsov2017semi}
Y.~Kuznietsov, J.~Stuckler, and B.~Leibe, ``Semi-supervised deep learning for
  monocular depth map prediction,'' in \emph{Proceedings of the IEEE conference
  on computer vision and pattern recognition}, 2017, pp. 6647--6655.

\bibitem{cho2019large}
J.~Cho, D.~Min, Y.~Kim, and K.~Sohn, ``A large rgb-d dataset for
  semi-supervised monocular depth estimation,'' \emph{arXiv preprint
  arXiv:1904.10230}, 2019.

\bibitem{sinz2004learning}
F.~H. Sinz, J.~Q. Candela, G.~H. Bak{\i}r, C.~E. Rasmussen, and M.~O. Franz,
  ``Learning depth from stereo,'' in \emph{Joint Pattern Recognition
  Symposium}.\hskip 1em plus 0.5em minus 0.4em\relax Springer, 2004, pp.
  245--252.

\bibitem{memisevic2011stereopsis}
R.~Memisevic and C.~Conrad, ``Stereopsis via deep learning,'' in \emph{NIPS
  Workshop on Deep Learning}, vol.~1, 2011, p.~2.

\bibitem{hoiem2005automatic}
D.~Hoiem, A.~A. Efros, and M.~Hebert, ``Automatic photo pop-up,'' in \emph{ACM
  SIGGRAPH 2005 Papers}, 2005, pp. 577--584.

\bibitem{saxena2005learning}
A.~Saxena, S.~H. Chung, A.~Y. Ng, \emph{et~al.}, ``Learning depth from single
  monocular images,'' in \emph{NIPS}, vol.~18, 2005, pp. 1--8.

\bibitem{saxena2008make3d}
A.~Saxena, M.~Sun, and A.~Y. Ng, ``Make3d: Learning 3d scene structure from a
  single still image,'' \emph{IEEE transactions on pattern analysis and machine
  intelligence}, vol.~31, no.~5, pp. 824--840, 2008.

\bibitem{liu2010single}
B.~Liu, S.~Gould, and D.~Koller, ``Single image depth estimation from predicted
  semantic labels,'' in \emph{2010 IEEE Computer Society Conference on Computer
  Vision and Pattern Recognition}.\hskip 1em plus 0.5em minus 0.4em\relax IEEE,
  2010, pp. 1253--1260.

\bibitem{ladicky2014pulling}
L.~Ladicky, J.~Shi, and M.~Pollefeys, ``Pulling things out of perspective,'' in
  \emph{Proceedings of the IEEE conference on computer vision and pattern
  recognition}, 2014, pp. 89--96.

\bibitem{krizhevsky2012imagenet}
A.~Krizhevsky, I.~Sutskever, and G.~E. Hinton, ``Imagenet classification with
  deep convolutional neural networks,'' \emph{Advances in neural information
  processing systems}, vol.~25, pp. 1097--1105, 2012.

\bibitem{laina2016deeper}
I.~Laina, C.~Rupprecht, V.~Belagiannis, F.~Tombari, and N.~Navab, ``Deeper
  depth prediction with fully convolutional residual networks,'' in \emph{2016
  Fourth international conference on 3D vision (3DV)}.\hskip 1em plus 0.5em
  minus 0.4em\relax IEEE, 2016, pp. 239--248.

\bibitem{li2015depth}
B.~Li, C.~Shen, Y.~Dai, A.~Van Den~Hengel, and M.~He, ``Depth and surface
  normal estimation from monocular images using regression on deep features and
  hierarchical crfs,'' in \emph{Proceedings of the IEEE conference on computer
  vision and pattern recognition}, 2015, pp. 1119--1127.

\bibitem{yin2019enforcing}
W.~Yin, Y.~Liu, C.~Shen, and Y.~Yan, ``Enforcing geometric constraints of
  virtual normal for depth prediction,'' in \emph{Proceedings of the IEEE/CVF
  International Conference on Computer Vision}, 2019, pp. 5684--5693.

\bibitem{huynh2020guiding}
L.~Huynh, P.~Nguyen-Ha, J.~Matas, E.~Rahtu, and J.~Heikkila, ``Guiding
  monocular depth estimation using depth-attention volume,'' \emph{arXiv
  preprint arXiv:2004.02760}, 2020.

\bibitem{bhat2021adabins}
S.~F. Bhat, I.~Alhashim, and P.~Wonka, ``Adabins: Depth estimation using
  adaptive bins,'' in \emph{Proceedings of the IEEE/CVF Conference on Computer
  Vision and Pattern Recognition}, 2021, pp. 4009--4018.

\bibitem{song2021monocular}
M.~Song, S.~Lim, and W.~Kim, ``Monocular depth estimation using laplacian
  pyramid-based depth residuals,'' \emph{IEEE Transactions on Circuits and
  Systems for Video Technology}, 2021.

\bibitem{xie2016deep3d}
J.~Xie, R.~Girshick, and A.~Farhadi, ``Deep3d: Fully automatic 2d-to-3d video
  conversion with deep convolutional neural networks,'' in \emph{European
  conference on computer vision}.\hskip 1em plus 0.5em minus 0.4em\relax
  Springer, 2016, pp. 842--857.

\bibitem{tosi2019learning}
F.~Tosi, F.~Aleotti, M.~Poggi, and S.~Mattoccia, ``Learning monocular depth
  estimation infusing traditional stereo knowledge,'' in \emph{Proceedings of
  the IEEE/CVF Conference on Computer Vision and Pattern Recognition}, 2019,
  pp. 9799--9809.

\bibitem{amiri2019semi}
A.~J. Amiri, S.~Y. Loo, and H.~Zhang, ``Semi-supervised monocular depth
  estimation with left-right consistency using deep neural network,'' in
  \emph{2019 IEEE International Conference on Robotics and Biomimetics
  (ROBIO)}.\hskip 1em plus 0.5em minus 0.4em\relax IEEE, 2019, pp. 602--607.

\bibitem{dai2019mvs2}
Y.~Dai, Z.~Zhu, Z.~Rao, and B.~Li, ``Mvs2: Deep unsupervised multi-view stereo
  with multi-view symmetry,'' in \emph{2019 International Conference on 3D
  Vision (3DV)}.\hskip 1em plus 0.5em minus 0.4em\relax IEEE, 2019, pp. 1--8.

\bibitem{ummenhofer2017demon}
B.~Ummenhofer, H.~Zhou, J.~Uhrig, N.~Mayer, E.~Ilg, A.~Dosovitskiy, and
  T.~Brox, ``Demon: Depth and motion network for learning monocular stereo,''
  in \emph{Proceedings of the IEEE Conference on Computer Vision and Pattern
  Recognition}, 2017, pp. 5038--5047.

\bibitem{hinton2015distilling}
G.~Hinton, O.~Vinyals, and J.~Dean, ``Distilling the knowledge in a neural
  network,'' \emph{arXiv preprint arXiv:1503.02531}, 2015.

\bibitem{romero2014fitnets}
A.~Romero, N.~Ballas, S.~E. Kahou, A.~Chassang, C.~Gatta, and Y.~Bengio,
  ``Fitnets: Hints for thin deep nets,'' \emph{arXiv preprint arXiv:1412.6550},
  2014.

\bibitem{peng2019correlation}
B.~Peng, X.~Jin, J.~Liu, D.~Li, Y.~Wu, Y.~Liu, S.~Zhou, and Z.~Zhang,
  ``Correlation congruence for knowledge distillation,'' in \emph{Proceedings
  of the IEEE/CVF International Conference on Computer Vision}, 2019, pp.
  5007--5016.

\bibitem{tung2019similarity}
F.~Tung and G.~Mori, ``Similarity-preserving knowledge distillation,'' in
  \emph{Proceedings of the IEEE/CVF International Conference on Computer
  Vision}, 2019, pp. 1365--1374.

\bibitem{yim2017gift}
J.~Yim, D.~Joo, J.~Bae, and J.~Kim, ``A gift from knowledge distillation: Fast
  optimization, network minimization and transfer learning,'' in
  \emph{Proceedings of the IEEE Conference on Computer Vision and Pattern
  Recognition}, 2017, pp. 4133--4141.

\bibitem{guo2018learning}
X.~Guo, H.~Li, S.~Yi, J.~Ren, and X.~Wang, ``Learning monocular depth by
  distilling cross-domain stereo networks,'' in \emph{Proceedings of the
  European Conference on Computer Vision (ECCV)}, 2018, pp. 484--500.

\bibitem{sun2021learning}
B.~Sun, X.~Ye, B.~Li, H.~Li, Z.~Wang, and R.~Xu, ``Learning scene structure
  guidance via cross-task knowledge transfer for single depth
  super-resolution,'' \emph{arXiv preprint arXiv:2103.12955}, 2021.

\bibitem{eigen2015predicting}
D.~Eigen and R.~Fergus, ``Predicting depth, surface normals and semantic labels
  with a common multi-scale convolutional architecture,'' in \emph{Proceedings
  of the IEEE international conference on computer vision}, 2015, pp.
  2650--2658.

\bibitem{yang2018deep}
N.~Yang, R.~Wang, J.~Stuckler, and D.~Cremers, ``Deep virtual stereo odometry:
  Leveraging deep depth prediction for monocular direct sparse odometry,'' in
  \emph{Proceedings of the European Conference on Computer Vision (ECCV)},
  2018, pp. 817--833.

\bibitem{tan2019efficientnet}
M.~Tan and Q.~Le, ``Efficientnet: Rethinking model scaling for convolutional
  neural networks,'' in \emph{International Conference on Machine
  Learning}.\hskip 1em plus 0.5em minus 0.4em\relax PMLR, 2019, pp. 6105--6114.

\bibitem{dosovitskiy2020image}
A.~Dosovitskiy, L.~Beyer, A.~Kolesnikov, D.~Weissenborn, X.~Zhai,
  T.~Unterthiner, M.~Dehghani, M.~Minderer, G.~Heigold, S.~Gelly,
  \emph{et~al.}, ``An image is worth 16x16 words: Transformers for image
  recognition at scale,'' \emph{arXiv preprint arXiv:2010.11929}, 2020.

\bibitem{fan2017point}
H.~Fan, H.~Su, and L.~J. Guibas, ``A point set generation network for 3d object
  reconstruction from a single image,'' in \emph{Proceedings of the IEEE
  conference on computer vision and pattern recognition}, 2017, pp. 605--613.

\bibitem{geiger2013vision}
A.~Geiger, P.~Lenz, C.~Stiller, and R.~Urtasun, ``Vision meets robotics: The
  kitti dataset,'' \emph{The International Journal of Robotics Research},
  vol.~32, no.~11, pp. 1231--1237, 2013.

\bibitem{liu2015learning}
F.~Liu, C.~Shen, G.~Lin, and I.~Reid, ``Learning depth from single monocular
  images using deep convolutional neural fields,'' \emph{IEEE transactions on
  pattern analysis and machine intelligence}, vol.~38, no.~10, pp. 2024--2039,
  2015.

\bibitem{gan2018monocular}
Y.~Gan, X.~Xu, W.~Sun, and L.~Lin, ``Monocular depth estimation with affinity,
  vertical pooling, and label enhancement,'' in \emph{Proceedings of the
  European Conference on Computer Vision (ECCV)}, 2018, pp. 224--239.

\bibitem{Ranftl2020}
R.~Ranftl, K.~Lasinger, D.~Hafner, K.~Schindler, and V.~Koltun, ``Towards
  robust monocular depth estimation: Mixing datasets for zero-shot
  cross-dataset transfer,'' \emph{IEEE Transactions on Pattern Analysis and
  Machine Intelligence (TPAMI)}, 2020.

\bibitem{godard2019digging}
C.~Godard, O.~Mac~Aodha, M.~Firman, and G.~J. Brostow, ``Digging into
  self-supervised monocular depth estimation,'' in \emph{Proceedings of the
  IEEE/CVF International Conference on Computer Vision}, 2019, pp. 3828--3838.

\bibitem{guizilini2020semantically}
V.~Guizilini, R.~Hou, J.~Li, R.~Ambrus, and A.~Gaidon, ``Semantically-guided
  representation learning for self-supervised monocular depth,'' \emph{arXiv
  preprint arXiv:2002.12319}, 2020.

\bibitem{guizilini20203d}
V.~Guizilini, R.~Ambrus, S.~Pillai, A.~Raventos, and A.~Gaidon, ``3d packing
  for self-supervised monocular depth estimation,'' in \emph{Proceedings of the
  IEEE/CVF Conference on Computer Vision and Pattern Recognition}, 2020, pp.
  2485--2494.

\bibitem{hur2020self}
J.~Hur and S.~Roth, ``Self-supervised monocular scene flow estimation,'' in
  \emph{Proceedings of the IEEE/CVF Conference on Computer Vision and Pattern
  Recognition}, 2020, pp. 7396--7405.

\bibitem{shu2020feature}
C.~Shu, K.~Yu, Z.~Duan, and K.~Yang, ``Feature-metric loss for self-supervised
  learning of depth and egomotion,'' in \emph{European Conference on Computer
  Vision}.\hskip 1em plus 0.5em minus 0.4em\relax Springer, 2020, pp. 572--588.

\bibitem{watson2021temporal}
J.~Watson, O.~Mac~Aodha, V.~Prisacariu, G.~Brostow, and M.~Firman, ``The
  temporal opportunist: Self-supervised multi-frame monocular depth,''
  \emph{arXiv preprint arXiv:2104.14540}, 2021.

\bibitem{silberman2012indoor}
N.~Silberman, D.~Hoiem, P.~Kohli, and R.~Fergus, ``Indoor segmentation and
  support inference from rgbd images,'' in \emph{European conference on
  computer vision}.\hskip 1em plus 0.5em minus 0.4em\relax Springer, 2012, pp.
  746--760.

\bibitem{chen2019structure}
X.~Chen, X.~Chen, and Z.-J. Zha, ``Structure-aware residual pyramid network for
  monocular depth estimation,'' \emph{arXiv preprint arXiv:1907.06023}, 2019.

\bibitem{kundu2018adadepth}
J.~N. Kundu, P.~K. Uppala, A.~Pahuja, and R.~V. Babu, ``Adadepth: Unsupervised
  content congruent adaptation for depth estimation,'' in \emph{Proceedings of
  the IEEE conference on computer vision and pattern recognition}, 2018, pp.
  2656--2665.

\bibitem{tian2019semi}
H.~Tian and F.~Li, ``Semi-supervised depth estimation from a single image based
  on confidence learning,'' in \emph{ICASSP 2019-2019 IEEE International
  Conference on Acoustics, Speech and Signal Processing (ICASSP)}.\hskip 1em
  plus 0.5em minus 0.4em\relax IEEE, 2019, pp. 8573--8577.

\bibitem{ji2019semi}
R.~Ji, K.~Li, Y.~Wang, X.~Sun, F.~Guo, X.~Guo, Y.~Wu, F.~Huang, and J.~Luo,
  ``Semi-supervised adversarial monocular depth estimation,'' \emph{IEEE
  transactions on pattern analysis and machine intelligence}, vol.~42, no.~10,
  pp. 2410--2422, 2019.

\bibitem{ramamonjisoa2019sharpnet}
M.~Ramamonjisoa and V.~Lepetit, ``Sharpnet: Fast and accurate recovery of
  occluding contours in monocular depth estimation,'' in \emph{Proceedings of
  the IEEE/CVF International Conference on Computer Vision Workshops}, 2019,
  pp. 0--0.

\bibitem{lee2019big}
J.~H. Lee, M.-K. Han, D.~W. Ko, and I.~H. Suh, ``From big to small: Multi-scale
  local planar guidance for monocular depth estimation,'' \emph{arXiv preprint
  arXiv:1907.10326}, 2019.

\bibitem{garg2016unsupervised}
R.~Garg, V.~K. Bg, G.~Carneiro, and I.~Reid, ``Unsupervised cnn for single view
  depth estimation: Geometry to the rescue,'' in \emph{European conference on
  computer vision}.\hskip 1em plus 0.5em minus 0.4em\relax Springer, 2016, pp.
  740--756.

\bibitem{paszke2019pytorch}
A.~Paszke, S.~Gross, F.~Massa, A.~Lerer, J.~Bradbury, G.~Chanan, T.~Killeen,
  Z.~Lin, N.~Gimelshein, L.~Antiga, \emph{et~al.}, ``Pytorch: An imperative
  style, high-performance deep learning library,'' \emph{arXiv preprint
  arXiv:1912.01703}, 2019.

\bibitem{loshchilov2019decoupled}
I.~Loshchilov and F.~Hutter, ``Decoupled weight decay regularization. arxiv,''
  \emph{Preprint published January}, vol.~4, 2019.

\bibitem{smith2018super}
L.~N. Smith and N.~Topin, ``Super-convergence: Very fast training of residual
  networks using large learning rates,'' 2018.

\end{thebibliography}

\bibliographystyle{IEEEtran}


%

\begin{IEEEbiography}{Michael Shell}
Biography text here.
\end{IEEEbiography}

\begin{IEEEbiographynophoto}{John Doe}
Biography text here.
\end{IEEEbiographynophoto}


\begin{IEEEbiographynophoto}{Jane Doe}
Biography text here.
\end{IEEEbiographynophoto}

\end{document}